\documentclass[runningheads]{llncs}
\usepackage{graphicx}
\usepackage{comment}
\usepackage{amsmath,amssymb} % define this before the line numbering.
\usepackage{color}
\usepackage{algorithm}
\usepackage{algorithmic}
\usepackage{subfigure}
\usepackage{booktabs}

\usepackage[width=122mm,left=12mm,paperwidth=146mm,height=193mm,top=12mm,paperheight=217mm]{geometry}

\def\eg{\emph{e.g}} 
\def\ie{\emph{i.e}}

\def\etal{\emph{et al}}

\begin{document}
% \linenumbers
\pagestyle{headings}
\mainmatter

\title{Powering One-shot Topological NAS with Stabilized Share-parameter Proxy} % Replace with your title

\titlerunning{ST-NAS}

\author{Ronghao Guo\inst{1} \and
Chen Lin\inst{2} \and
Chuming Li\inst{2} \and
Keyu Tian\inst{1} \and \\ 
Ming Sun\inst{2} \and
Lu Sheng\inst{1} \and
Junjie Yan\inst{2}
}

\authorrunning{R. Guo, C Lin et al.}
\institute{
  Beihang University 
  \email{\{16211042,17375491,lsheng\}@buaa.edu.cn} \and
  SenseTime Research 
  \email{\{linchen,lichuming,sunming1,\\yanjunjie\}@sensetime.com}
}
% \end{comment}
%******************
\maketitle

\begin{abstract}

% One-shot searching of network architectures has attracted much interest from the research community of neural architecture search, due to its remarkable training efficiency and capacity to discover high performance models. 
One-shot NAS method has attracted much interest from the research community due to its remarkable training efficiency and capacity to discover high performance models.
However, the search spaces of previous one-shot based works usually relied on hand-craft design and were short for flexibility on the network topology.
In this work, we try to enhance the one-shot NAS by exploring high-performing network architectures in our large-scale Topology Augmented Search Space (\ie, over $3.4 \times 10^{10}$ different topological structures).
Specifically, the difficulties for architecture searching in such a complex space has been eliminated by the proposed stabilized share-parameter proxy, which employs Stochastic Gradient Langevin Dynamics to enable fast shared parameter sampling, so as to achieve stabilized measurement of architecture performance even in search space with complex topological structures.
The proposed method, namely Stablized Topological Neural Architecture Search (ST-NAS), achieves state-of-the-art performance under Multiply-Adds (MAdds) constraint on ImageNet.
Our lite model ST-NAS-A achieves $76.4\%$ top-1 accuracy with only $326$M MAdds.
Our moderate model ST-NAS-B achieves $77.9\%$ top-1 accuracy just required $503$M MAdds.
Both of our models offer superior performances in comparison to other concurrent works on one-shot NAS.
% \grh{[173 / 150 words limit]}

% One-shot based method has achieved great success in NAS area due to its training efficiency and ability to discover high-performing models. 
% However, the search space of previous works are usually carefully designed and do not enjoy too much flexibility on network topology.

% In this work, we try to explore high-performing network architectures in our Topology Augmented Search Space using one-shot NAS.  
% The proposed search space contains over $3.4\times 10^{10}$ different network topology. 
% The complex search space introduces extra difficulties for NAS.
% Specifically, we show that the performance estimation using shared parameters becomes noisy when the search space contains complex topology.
% In order to eliminate the noise, we propose to measure performance via estimating the expectation among multiple runs.
% A fast shared parameters sampling methods based on Stochastic Gradient Langevin Dynamics is developed to approximate shared weights obtained in different runs.

% The resulted method, namely Stabilized Topological Neural Architecture Search (ST-NAS), achieves state-of-the-art performance under Multiply-Adds (MAdds) constraint on ImageNet. Our lite model ST-NAS-A achieves 76.4\% top-1 accuracy with only 326M MAdds. Our moderate model ST-NAS-B achieves 77.9\% top-1 accuracy with around 503M MAdds. Both of our models achieve superior performance compared with other works on one-shot NAS.

\end{abstract}

\section{Introduction}

% ---main-pic---
\begin{figure}[t]
    \begin{minipage}[h]{0.48\textwidth}
      \centering
      \includegraphics[width=1\textwidth]{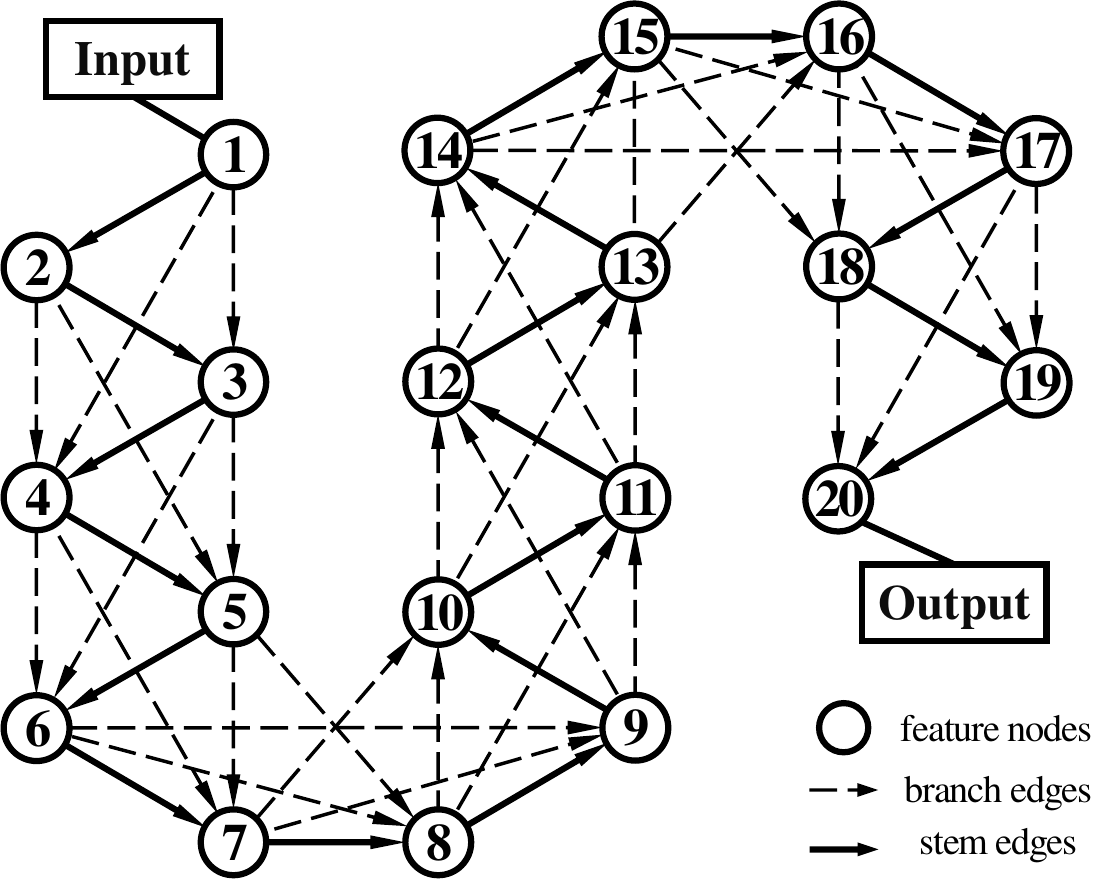}
      \caption{An instance of our topology augmented search space. It contains over $3.4\times 10^{10}$ different network topologies which enables us to explore complex network typologies. Solid line denotes the chain-structured stem edges, and dotted line represents branch edges which connects feature maps with depth difference 2 or 3.}
      \label{fig:main-pic}
    \end{minipage}
    \hfill
    \begin{minipage}[h]{0.44\textwidth}
        \centering
        \subfigure[SPOS space]{
        \begin{minipage}[t]{0.45\textwidth}
            \centering
            \includegraphics[width=0.95\textwidth]{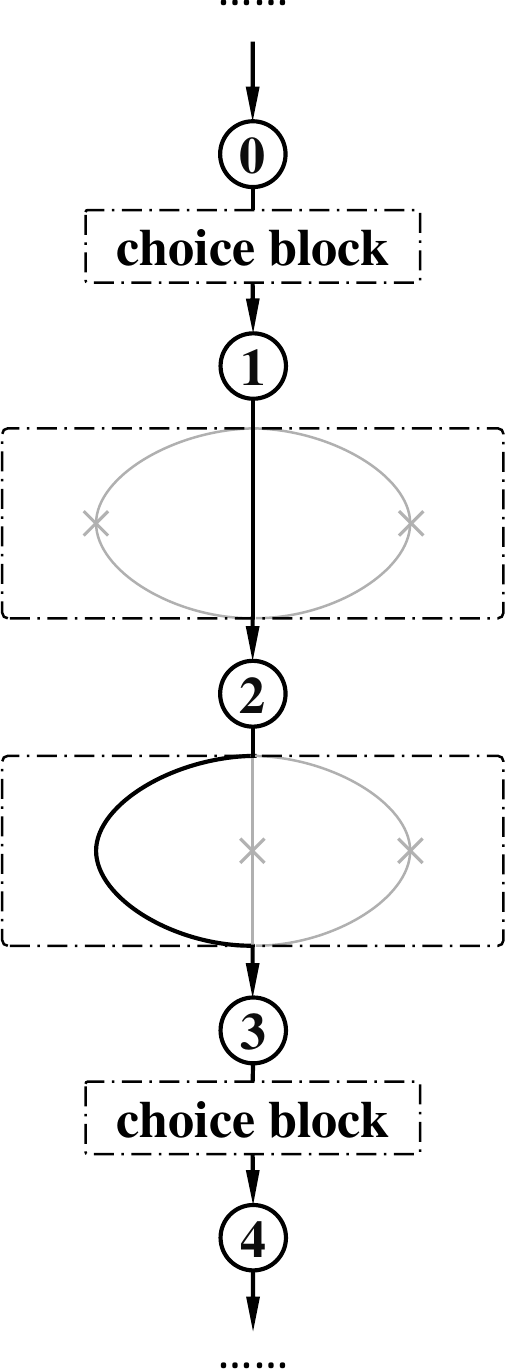}   
        \end{minipage}
        }
        % \hfill
        \subfigure[DARTS space]{
        \begin{minipage}[t]{0.45\textwidth}
            \centering
            \includegraphics[width=0.95\textwidth]{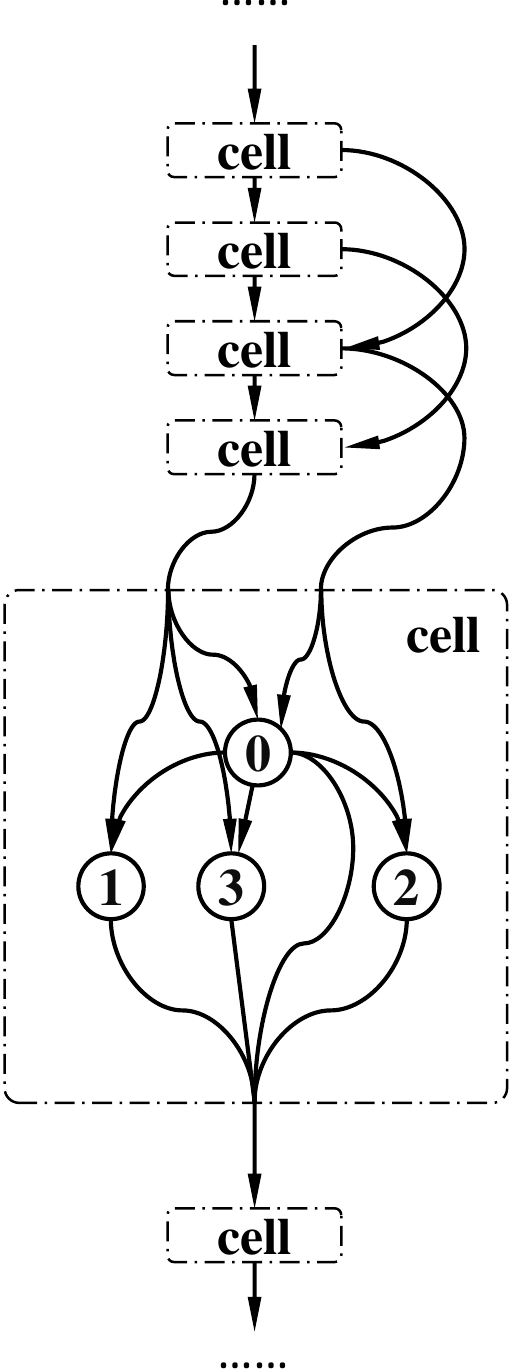}
        \end{minipage}
        }
        \caption{Two typical search spaces in previous work. Fig. (a) shows a chain-structured space while Fig. (b) shows a cell-based search space. }
        \label{fig:two-space}
    \end{minipage}
\end{figure}

Significant progress made by convolution neural networks (CNN) in challenging computer vision tasks has raised the demand to design powerful neural networks. 
Instead of manually design, Neural architecture search (NAS) has demonstrated great potentials in recent years.
Early works of NAS by Real~\etal\cite{real2017large,real2019regularized} and Elsken~\etal\cite{elsken2017simple}
% Early works of NAS (Real \etal\cite{real2017large,real2019regularized}; Elsken \etal\cite{elsken2017simple}, 2017)
%
achieved promising results but can only be applied to small datasets due to their large computation expenses.
To this end, one-shot based methods have drawn much interest thanks to its promising training efficiency and remarkable ability to discover high-performing models. One-shot method usually utilizes a hyper-network, which subsumes all architectures in the search space, and use shared weights to evaluate different architectures. 
%
% \grh{[Add more details about weight sharing]} %One-shot NAS usually takes use of weight sharing strategy to train a single large network.}
% question1
%Generally, weight sharing  

However, the search space of previous works (\eg, shown in Fig. \ref{fig:two-space}) were usually carefully designed and did not enjoy too much flexibility on the network topology. %, shown in Fig. \ref{fig:two-space}.
% Architecture search space is crucial to the performance of a NAS method.
% Large search space probably contain high-performing models while increase the difficulty of searching.
%
For example, as one of mostly applied search spaces in the one-shot literature, the chain-structured search space\cite{guo2019single} has sequentially connected intermediate feature maps, between which the edges are chosen from a set of computation operations.
Networks with better operations can be discovered on this search space, but the network topology remains trivial.
However, previous works \cite{du2019spinenet,huang2017densely} on network architecture design proved that complex topology will tremendously enhance the performance of deep learning models.
%
% We argue that ** topological diversity 
We argue that complex topological structures added in search space will improve the performance of the searched network architectures as shown in Table~\ref{tab:rand-skip}.
%, since \grh{it can combine high-level and low-level features. [Why need more topological structures? It is important to amateur readers.]}
% Question: 拓扑多样性优点 融合高次、低次特征？
% For example, a chain-structured search space from\cite{guo2019single} is mostly used in the one-shot literature.
% %
% In chain-structured search space, intermediate feature maps are connected sequentially, and the edges between them have multiple choices of computation operations.
% NAS methods operating on this search space would discover networks with better operation selection but trivial topology .

In this work, we are interested in exploring complex network typologies with one-shot method. We propose a novel network architecture search space shown in Fig. \ref{fig:main-pic} which contains over $3.4\times 10^{10}$ different network topologies, enabling the discovery of complex topology networks. The search space is obtained by introducing numerous computation modules as edges between nodes. A topology based architecture sampler is also introduced to sample architectures during one-shot training stage from the hyper-network. 
However, the great diversity introduced by topologies brings difficulties to the one-shot approach. Specifically, we observe high variance of performance estimation through the one-shot shared parameters in two cases: estimation through shared parameters at different epochs of a single run and estimation through shared parameters obtain in different runs. Zhang \etal \cite{zhang2020deeper} explore the reason behind the variance of ranking under weight sharing strategy. Thus the ranking ability of shared parameters is compromised. 

% However, we observed that the complex topology architectures in hyper-network introduces noise in the weight distribution which

% (evaluating the search phase, pc-nas).

% ere are two components in a NAS method: Architecture Search Space and Search Process. 

To eliminate the interference of complex topologies, we estimate the expectation of architecture performance in additional training epochs of hyper-network via multiple samples of shared parameters. An fast weights sampling methods based on Stochastic Gradient Langevin Dynamics is developed to sample shared parameters efficiently.

The resulted Stabilized Topological Neural Architecture Search (ST-NAS) achieves compatible performance with the state-of-the-art NAS method.
% at much lower search cost and outperforms previous one-shot methods, \grh{due to the topology enhancement and .}
% we can achieve equivalent performance with lower computation cost.
The resulted architecture ST-NAS-A obtains 76.4\% top-1 accuracy with only 326M MAdds.
A larger architecture ST-NAS-B obtains 77.9\% top-1 accuracy with around 503M MAdds.

To summarize, our main contributions are as follows:
\begin{enumerate}
\item We introduce a topology augmented neural architecture search space that enables the discovery of efficient architectures with complex topology.
\item To relieve the complex topology's interference on model ranking, we modified model evaluation based on the expectation of the sharing parameters' performance.
\item We empirically demonstrate improvements on ImageNet classification under the same MAdds constraints compared with previous work, and show that the searched architectures transfer well to COCO object detection. %\grh{[Should we claim COCO results here?]}
% \item Our methods achieves state-of-the-art performance 
% Question: COCO 结果需要作为contribution吗，我觉得可以不用；
\end{enumerate}

%------------------------------------------------------------------------
\section{Related Work}

% NAS method has improve CNN performance a lot in vision tasks. Early NAS works can be typically considered as an agent-based explore and exploit process. Such methods are pretty computationally expensive. Some recent works reduce computation cost via training a single hyper-network where the weights of the same operators in different models are shared. All candidate models are comprised in the hyper-network therefore. As the most representative one, NAS offers us a more creative and effective approach for discovering better-performed network architectures automatically.

%  and Guo \etal\cite{guo2019single} both propose a one-shot architecture search method by utilizing an over-parameterized network which comprises all the candidate paths sampled with uniform probabilities.

% Such methods have made significant progress in computation reduction, as well as enhance the performance of weight sharing to a higher level. However, one-shot method has shown 
% owing to coupled parameters
% , their topological diversities are partly limited, which may lead to insufficient searching.

%Two widely used agents are reinforcement learning agent \cite{zoph2016neural, zhong2018blockqnn, zoph2018learning, guo2019irlas, tan2019mnasnet, baker2016designing} and evolution agent \cite{real2017large, lu2018nsga, chen2018reinforced, guo2019single}. The network generated are usually trained from scratch on the original or proxy datasets to estimate its performance. These methods are facing problem of expensive computational cost. 
Recently, auto machine learning methods have received a lot of attention due to its ability to design augmentation \cite{cubuk2018autoaugment,lin2019online}, loss function \cite{li2019lfs} and network architectures \cite{zoph2016neural,real2019regularized,brock2017smash,zhong2018blockqnn,liu2018darts,guo2019single,guo2019irlas,liang2019computation,li2019improving}.

Early neural architecture search (NAS) works normally involves reinforcement learning \cite{baker2016designing,zoph2016neural,zhong2018blockqnn,zoph2018learning,guo2019irlas,tan2019mnasnet} or evolution algorithm \cite{real2017large,lu2018nsga} to search for high-performing network architectures. However, these methods are usually computationally expensive which limits its uses in real scenarios.

Recent attentions have been focused on alleviating the computation cost via weight sharing method. This method usually contains a single training process of an over parameterized hyper-network which subsumes all candidate models, \ie, weights of the same operators are shared across different sub-models. Notably, Liu \etal\cite{liu2018darts} proposes continuous relaxations which enables optimizing network architectures with gradient decent, 
Cai \etal\cite{cai2018proxylessnas} proposes a proxy-less method to search on target datasets directly and 
Bender \etal\cite{bender2018understanding} introduces one-shot method to decouple training and searching stages. Our NAS work take the use of the weight sharing hyper-network but relieve the variance during model training.

Early hand-craft neural networks \cite{he2016deep,szegedy2016rethinking,szegedy2017inception} tend to stack repeated motifs. Works in \cite{szegedy2015going,he2016deep,huang2017densely,hu2018squeeze} introduce different manual designed network topologies and result in performance gain. 
% achieve performance gain with the help of network topologies they proposed. 

Motivated by manual designed architectures \cite{he2016deep,szegedy2016rethinking,szegedy2017inception}, a widely used search space in works \cite{zoph2016neural,liu2018darts,zhong2018blockqnn,lu2018nsga,guo2019irlas} are proposed to search for such motifs, dubbed cells or blocks, rather than all possible architectures. This search space is called cell-based space. % in our work. 
Another widely used search space adopted in \cite{cai2018proxylessnas,guo2019single,tan2019mnasnet,xiong2019resource} is called chain-structured space. This space sequentially stacks several operation layers where each layer serves its output as next layer's input. NAS methods are adopted to search for operation layers in different position of this space.
Work in \cite{xie2019exploring} explores random wiring networks with less human prior and achieves comparable performance with manual designed networks. %Work in \cite{liang2019computation} use NAS method to improve detection performance on a small search space to reallocate the computation blocks in different resolutions.

\section{Approach}
\label{approach}

Methods for NAS usually consist of three basic components: search space, performance estimation and search strategy.
% In our work, we mainly focus on the search space and the performance estimation.
In this section, we first introduce our novel Topology Augmented Search Space and a new sampling strategy for hyper-network training in this particular space. 
Secondly, we provide new model performance estimation approach to relieve the variance of model ranks during the training of hyper-network.
Finally the evolution algorithm for network search is described.

\subsection{Topology Augmented Search Space}

\subsubsection{Motivation} 
To demonstrate the improvement of complex topology against a sequential structure, we take ResNet-18 as a baseline and shows a subtle change on the topology obtains obvious performance boost.
We randomly add 4 residual blocks to connect the feature maps of blocks in ResNet-18's \cite{he2016deep} chain structure with 3 random seeds, and rescale the width to keep the same FLOPs, the results are in Table \ref{tab:rand-skip}. The 3 
complex structures imply the great potential of topology-based structure search.

\begin{table}[b]
\begin{center}
\begin{tabular}{ccccc}
\toprule[1.5pt]

Architecture &Res-18 & Rand0 & Rand1 & Rand2\\
\midrule[1pt]
Accuracy (\%) & 70.2 & 71.5 & 72.0 & 69.6\\

\bottomrule[1.5pt]
\end{tabular}
\end{center}
\caption{The accuracy of ResNet-18 and three networks with 4 random skip residual blocks added on the baseline. The three networks are scaled to keep the FLOPs same with ResNet18. Obvious boosting is obtained via exploration in a more complex topology space.}
\label{tab:rand-skip}
\end{table}

\subsubsection{Search Space} 
A neural network is denoted as a directed acyclic graph (DAG) defined by $E,V$, where the node $v_i \in V$ indicates the feature connected by edge $e_i \in E$ and edges represent CNN operators. The nodes $v_1, v_2, ...$ are indexed by the order of computation of their corresponding feature maps.

In our formulation, each $e_i $ is a minimum search unit, also referred as a choice block, which contains a set of candidates computation blocks. A hyper-network is the network which subsumes all the sub-architectures in the search space.
% Each sub-architecture $m_k$ is derived by the numb choosing the candidate blocks in all edges as particular ones.
Following the previous works, we divide our search space into several sub DAGs (stages), each of which downsamples the input by a factor $2$.

%---------------------- Our Block Detailed Pic ----------------------

\begin{figure}[t]
    \begin{minipage}[t]{0.4\textwidth}
        \centering
        \includegraphics[width=4.0cm]{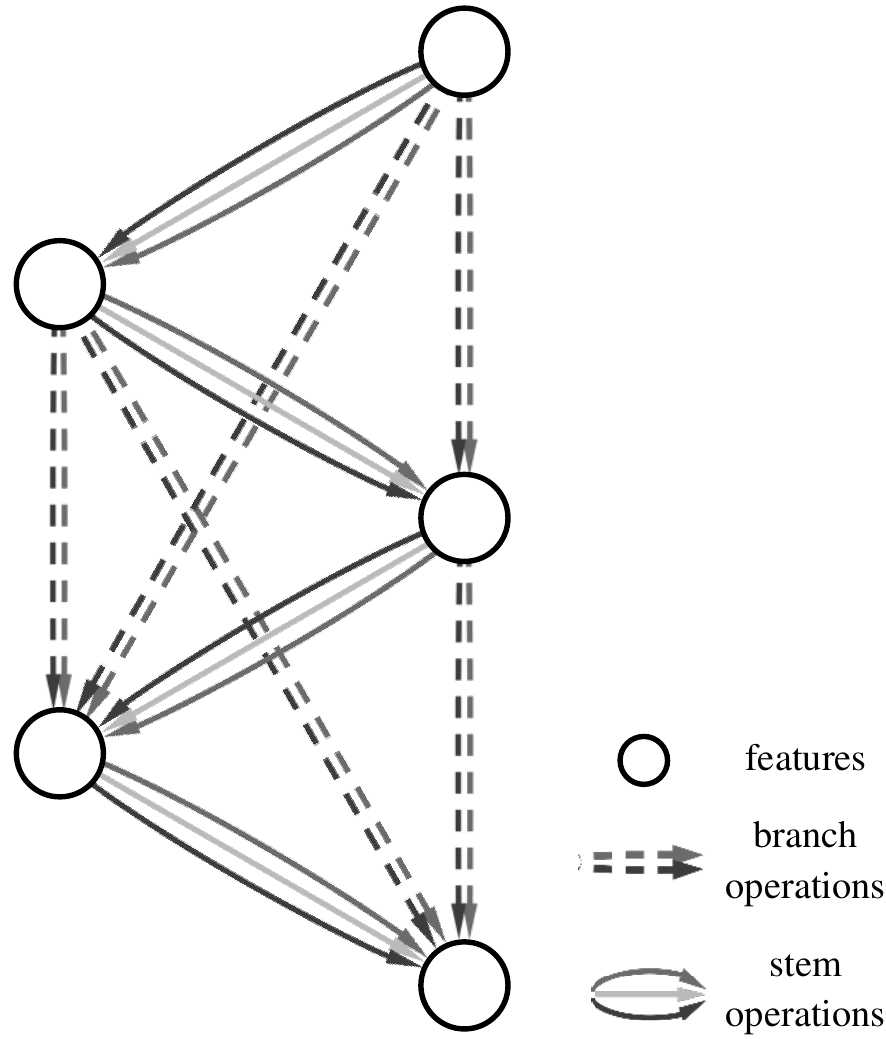}
        \caption{Illustration of candidates in stem edges and branch edges. }
        \label{fig:cell-pic}
    \end{minipage}
    \hfill
    \begin{minipage}[t]{0.55\textwidth}
        \centering
        \includegraphics[width=5.5cm]{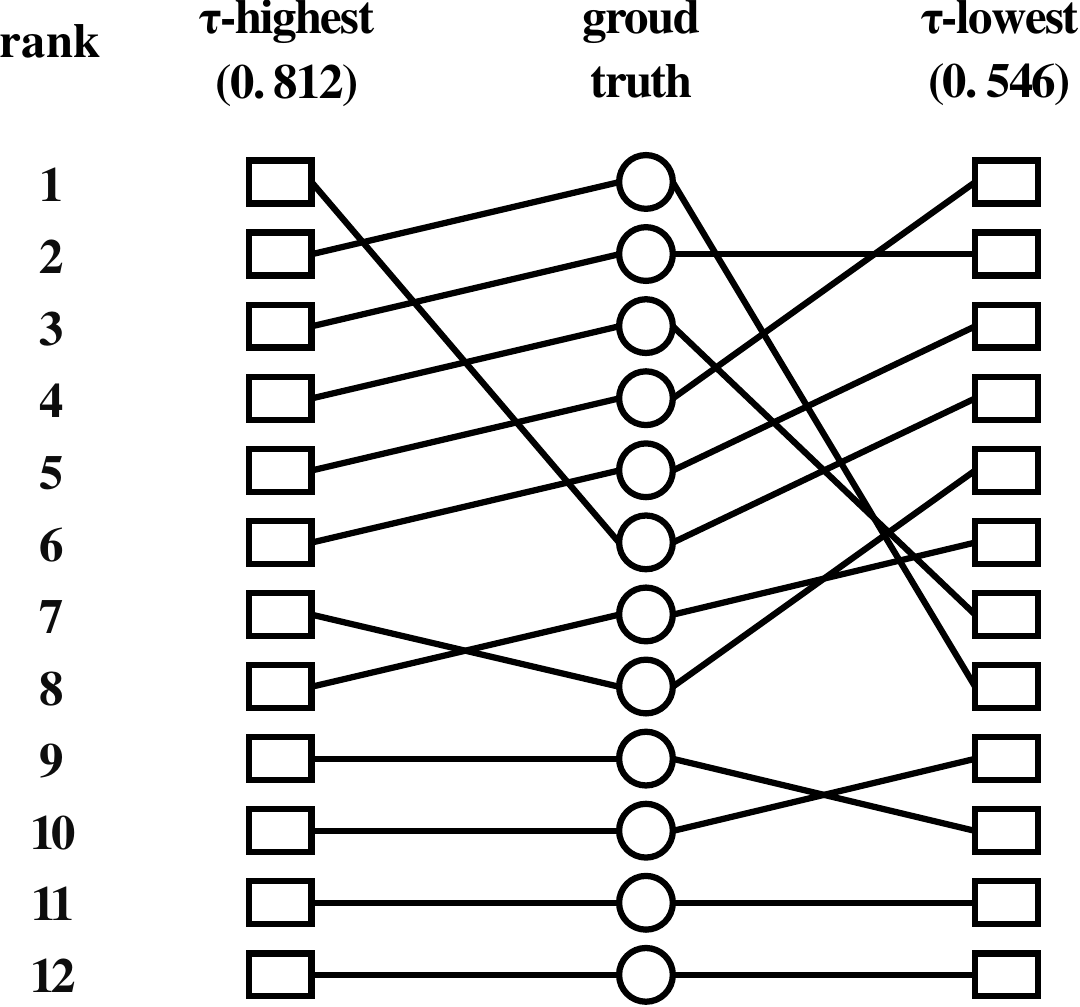}
        \caption{Detailed ranking of the best and the worst among 20 runs. The ground truth ranking is provided in the middle. This figure shows that the quality of ranking exceedingly differs from 
        s. }
        \label{fig:tau-runs}
    \end{minipage}
\end{figure}

To enable the discovery of complex topology architectures, a novel topology augmented search space is proposed.
In our search space, edges are divided into two categories, stem edges and branch edges, detailed in Fig. \ref{fig:cell-pic}.

\textbf{Stem Edges} are non-removable edges which always appear in candidate architectures.
The stem edges exist between all node pairs $(v_i, v_j)$, where $|i-j|=1$ . 
Stem edges are chain-structured, which sequentially connect all consecutive nodes in each stages.
We use the 9 kinds of linear bottlenecks (LB) \cite{sandler2018mobilenetv2} as the candidate choices of stem edges. 
Further, on stem edges between feature maps with the same resolution, identity operation is added as an extra candidate to enhance topological diversity and depth flexibility. Therefore, there are $10$ choices in the sequential structures. 

% The the output of branch edge are directly. 
\textbf{Branch Edges} are optional to contribute to topology diversity in the search space.  
The branch edges exist between all node pairs $(v_i, v_j)$, where $|i-j|\in \{2,3\}$. The candidate choices of branch edges are the same to stem edges. Differently, the branch edges could be abandoned flexibly.  

When $v_i$ and $v_j$ has different resolution, the stride of convolution operation in the edge is automatically adapted to align the feature maps.
The number of nodes in a single stage is required to define the search space. Based on previous method, we set the number of nodes in each stage as 2, 2, 4, 8, 4. 

The search space we proposed ensures network topology complexity. %while maintaining computational efficient. 
Network topology in this work is defined as the DAG formed by nodes and edges. 
For nodes, the total number of topology is $2^{2(n-3)+1}$. The search space we used in the experiment contains 20 nodes in total, which is around $3.4 \times 10^{10}$ topologies.
For comparison, the topologies contained in cell-based search space is around $7.2\times 10^9$. %\grh{micro or macro differences.}

\subsection{Training the One-shot Hyper-network}
\label{hyper-network_train}
One-shot method uses the hyper-network to estimate the performance of architectures.
Since huge amount of architectures exists in the hyper-network concurrently, training the hyper-network in whole will make the parameters of different architectures correlated with each other.
To reduce the correlation, one-shot method samples a new network architectures $m_k$ at each gradient step and update the only the activated part of the shared parameters.
\begin{equation}
\begin{split}
  \theta_T = \theta_{T-1} +\alpha \cdot \nabla L(\mathcal{F}(X, m_k, \theta_{T-1}), Y), \\
  m_k \sim P_t(m_k).
\end{split}
\end{equation}
$\mathcal{F}$ makes prediction of input $X$ utilizing sampled model $m_k$. Thus the gradient of parameters unused by $m_k$ remains zero.
The architecture sampling distribution $P_t$ is usually set to trivial uniform sampling \cite{guo2019single} across the choice for each single edge.
% \del{To make the training procedure more stable, we propose a new sampler which gradually transition from simple topology to complex topology. For the stem edges, we use the same uniform sampling strategy.}

Suppose there are $I_{stem}$ choices for stem edges and $I_{branch}$ for branch edges other than \textit{none}, a simple uniform sampling strategy in our search space can be described as:
% \begin{equations}
\begin{align}
p_{stem}(o_i) &= \frac{1}{I_{stem}},\\
% \end{equation}
% \begin{equation}
p_{branch}(o_i) &= \frac{1}{I_{branch}+1}.
\end{align}
% \end{equations}
%\del{ we set probability for the branch edge to be dropped as} 

However, the network sampled under this strategy in our space tends to sample architectures with high computational cost, because each of the large amount of branch edges has a low probability to be \textit{none}. Consequently, the architecture with low computational cost in the hyper-network will under-fit, which would cause a bias in evaluation stage.
Thus, the sampling strategy needs further consideration. 
The whole training process of hyper-network can be finded in Algo. \ref{algo:train_hyper-network}. 
Suppose that $C_{target}$ is our target MAdds and $C_{m_k}$ is the MAdds of architecture $m_k$, the sampling strategy should meet:

\begin{gather}
    E_{m_k \sim P_t(m_k)}[C_{m_{k}}] = C_{target}.
\end{gather}

To meet the constraints on expected computation, the sampling probability of \textit{none} choice in branch edges, $p_{drop}$, is defined to adjust the expected computation cost of sampled networks:

\begin{align}
p_{branch}(o_i) = 
\begin{cases}
\frac{1-p_{drop}}{I_{branch}}, ~ o_i \not= \text{\textit{none}}, \\
p_{drop}, ~ o_i     = \text{\textit{none}}.
\end{cases}
\end{align}

\begin{algorithm}
\caption{Hyper-network Training}
\label{algo:train_hyper-network}
\begin{algorithmic}[1]
\STATE \textbf{Inputs:} $D_{train}$, $T$, $B$
\STATE $G(E,V)$ = InitializeHyperNetwork()
\FOR{$t = 1:T$}
\STATE $p_{stem} = \frac{1}{I_{stem}}$
% \STATE $p_{drop} = 1 - \frac{t}{T(I_{branch}+1)}$
% \STATE $p_{branch} = \frac{1-p_{drop}}{I_{branch}}$
\STATE $p_{branch} = \frac{1-p_{drop}}{I_{branch}} ~ if ~ o_i ~ \not= none ~ else ~ p_{drop}$
\STATE $E_{stem}^{'}$   = Sample($E_{stem}$, $p_{stem}$)
\STATE $E_{branch}^{'}$ = Sample($E_{branch}$, $p_{branch}$)
\STATE $m$              = $G(E_{stem}^{'} \cup E_{branch}^{'}, V)$
\STATE $D_{batch}$      = Sample($D_{train}$, $uniform$, $B$)
\STATE TrainForOneStep($m$, $D_{batch}$)
\ENDFOR
\STATE \textbf{Outputs: } $G(E,V)$
\end{algorithmic}
\end{algorithm}

% \del{------------------------------------------------------------------------------------------}
% \grh{where $p_{drop}$ is used to adjust the expected computation cost of sampled networks.}
% \del{
% \begin{equation}
% p_{drop}=1-\frac{t}{T(I_{branch}+1)}
% \end{equation}
% }
% \del{Where $t$ is the current iteration, and $T$ is the total iteration of the hyper-network training.
% Then the probability of operators in branch edge is simply
% \begin{equation}
% p_{branch}(o_i) = \frac{1-p_{drop}}{I_{branch}}
% \end{equation}
% }
% The imposed sampling distribution gradually transitions from chain-structured architectures to complex architectures.
% The details of the hyper-parameters we used for hyper-network training could be find in \ref{sec:exp-setting}.

\subsection{Stabilizing Performance Estimation}

In search stage, evaluating an architecture through the shared parameters is essential for exploring promising results.
Previous work on one-shot method usually measure the network performance with fully trained hyper-network weights directly.
In this section, we first demonstrate our observation on random shuffling of candidates architectures ranking in our search space.
Then we introduce our approach to improve the ranking stability. 

% \begin{figure}
% \begin{center}
% \centering
%     \includegraphics[width=1\linewidth]{pic/te-rank.pdf}
% \end{center}
% \caption{The ranks of 10 random architectures during the last 20 training epochs, where drastic fluctuation can be observed. Ranks of three architectures at two time steps are shown in the figure, and each of them has different ranks at the two steps.}
% \label{fig:training_rank}
% \end{figure}

\subsubsection{Instability of One-shot NAS}

Since the hyper-network is trained $T$ iterations, the shared parameter obtained after training is denoted as $\theta_{T}$. 
We define a accuracy function $Acc(m_{k}, \theta)$ which maps the model architecture $m_k$ and hyper-network weights $\theta$ to the validation set accuracy.
The value of $Acc$ function can be estimated by simply loading the weight used by $m_k$ and testing the model performance on validation set.
The score function, denoted as $S(m_k)$,  of previous approach is simply
\begin{equation}
 S(m_k)=Acc(m_k, \theta_T).
\end{equation}
However, the true score function should be the actual performance of the model $m_k$ on validation set: $Acc(m_k, \theta_{m_k})$, where  $\theta_{m_k}$ denotes the weight obtained by sampling and training $m_k$ only.
One-shot approach takes a approximation to reuse the shared parameters for different architectures.
Although this is empirically useful, we observe high variance of the model ranking in two cases: rankings at different epochs and rankings by different runs.

\begin{figure}[t]
\begin{center}
    \includegraphics[width=6.8cm]{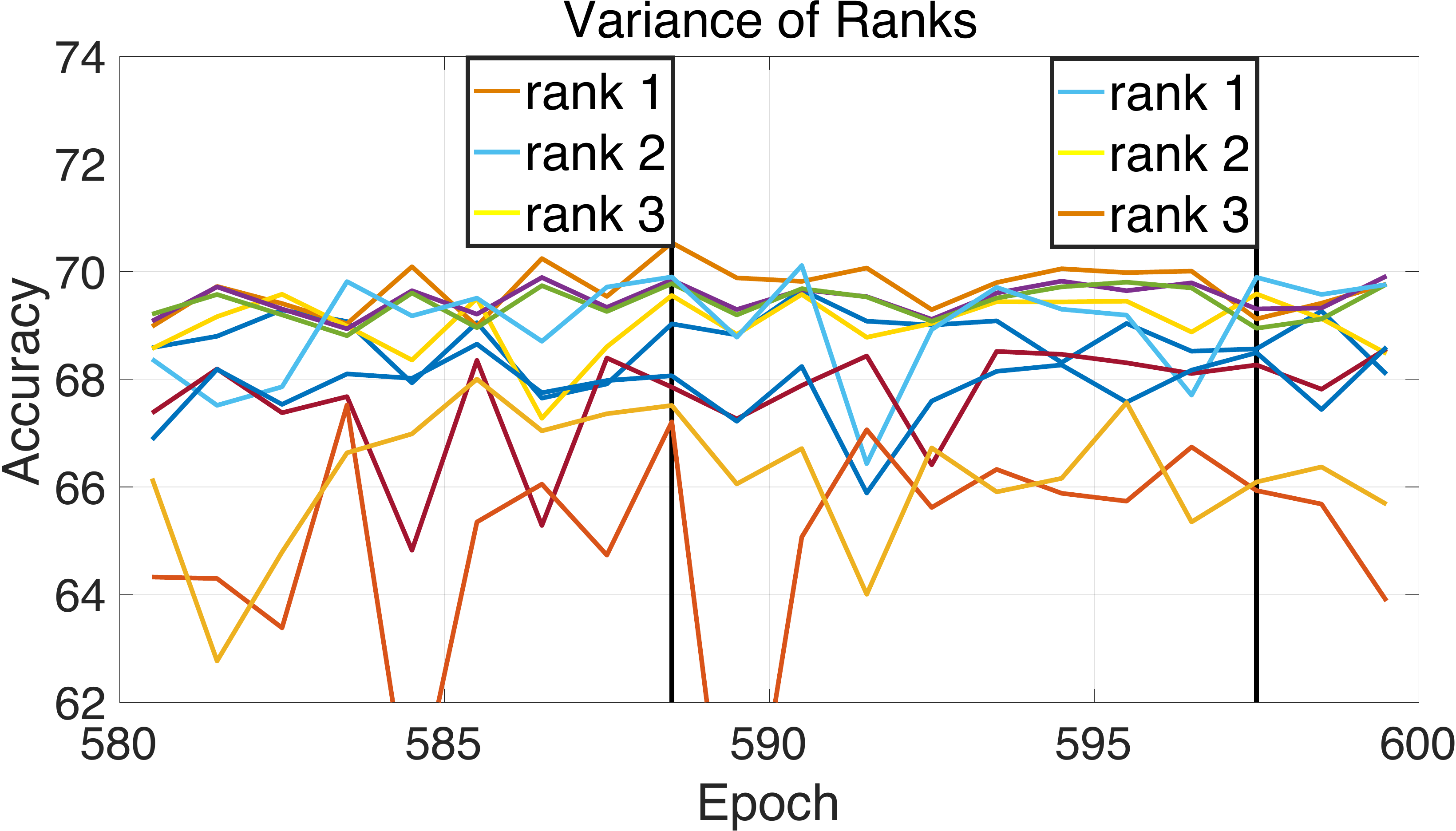}
\end{center}
   \caption{The ranks of 10 random architectures during the last 20 training epochs, where drastic fluctuation can be observed. Ranks of three architectures at two time steps are shown in the figure, and each of them has different ranks at the two steps.}
\label{fig:training_rank}
% \label{fig:tau-runs}
\end{figure}

We  randomly  sample  a  set  of  architecture  and  obtain their independent weight. We  rank  their performance under shared parameters on validation set by different checkpoints at the last 20 epochs training of hyper-network. As shown in Fig. \ref{fig:training_rank} , the rank of a single checkpoint fluctuates a lot during hyper-network training process and hardly distinguishes the performance of architectures.
If we repeat the hyper-network training with different random seeds for 20 times and obtain shared parameters $\theta_{T}^i, i\in[1, 20]$.  We quantify the correlation between rank of each  $\theta_{T}^i$ and ground truth rank by Kendall's $\tau$ coefficient \cite{kendall1938new}. Here, we show the ranking performance of the best and worst runs in Fig. \ref{fig:tau-runs}.

These two observations imply the necessity of a stabilized evaluation strategy.
To present our strategy, formulation of the instability need to be introduced.
In this paper, we model the performance estimation randomness as an unbiased noise. 
Since the shared parameters is fundamentally different from the parameters trained independently, we use a function $\phi$ to model the affect of weight sharing.
General consensus has been reached: empirical $Acc(m_k,\theta_{T})$ provides inaccurate but useful ranking, which demonstrates the desired rank preserving 
effect of $\phi$.
In summary, our model to describe the quantity relationship is:
\begin{align}
Acc(m_k, \theta_T) &= \phi(Acc(m_k, \theta_{m_k})) + \upsilon ,\\
E[\upsilon] &= 0.
\end{align}

It is obvious that the existence of the noise term $\upsilon$ would hurt the model ranking.
The most trivial approach to alleviate the negative effects of $\upsilon$ is to train multiple hyper-networks, and eliminate the noise by taking expectation.
However, this approach requires several times more computation resources for hyper-network training.

\subsubsection{SG-MCMC Sampling}
\label{sec:mcmc}
The sampling process is described in Algo. \ref{algo:mcmc-sampling}. In order to obtain high-quality low correlation samples of optimized shared parameters efficiently, we investigate the rich literature of Markov Chain Monte Carlo (MCMC) sampling methods \cite{bishop2006pattern}.
Recently, a few works demonstrate that constant learning rate stochastic gradient decent could be modified to Stochastic Gradient Langevin Dynamics (SGLD) to realize a Stochastic Gradient MCMC method under mild assumption\cite{chen2016bridging,welling2011bayesian}. Here, we apply SGLD \cite{welling2011bayesian,teh2016consistency} to approximate iid samples of share parameters posterior.
The update rule we use, is simply
\begin{equation}
\label{SGLD_update}
\begin{split}
\Delta\theta_{T} = \alpha \nabla (\frac{1}{\mathcal{B}} \sum_i^{\mathcal{B}} &L(\mathcal{F}(x_i, m_t, \theta_{T}), y_i) ) +  \sqrt{2\alpha }~\epsilon, \\
m_t &\sim P_t(m) \text{ and } \epsilon \sim \mathcal{N}(\boldsymbol{0},I).
\end{split}
\end{equation}
Here $\mathcal{B}$ is the number of data used to compute gradients (batch size). The step size $\alpha$ is set to the final learning rate of sub-net training. To ensure the independence, we generate each sample after SGLD update iterates for a data epoch.

To generate the iid samples of shared weights, we load the weights $\theta_{T}$ of the hyper-network after its training finishes, and set $\theta_{T}^0=\theta_{T}$ as the initial sample. Then, for each $\theta_T^i$, we apply SGLD to obtain the next sample of parameter posterior $\theta_T^{i+1}$ with the rule in Eq. (\ref{SGLD_update}). Thus we can obtain multiple samples of hyper-network parameters. %Once we have the SGLD samples, we fine-tune them with standard SGD with momentum to reach the convergence under aforementioned hyper-network training procedure in Section \ref{hyper-network_train}.

\begin{algorithm}[t]
\caption{Shared Parameter Sampling by SGLD}
\label{algo:mcmc-sampling}
\begin{algorithmic}[1]
\STATE \textbf{Inputs:} $D_{train}$, $T_{SGLD}$, $T_{epoch}$, $B$, $G(E,V)$, $\alpha$
\STATE $\mathcal{G}_{sample}$ = $\emptyset$
\FOR{$t = 1:T_{SGLD}$}
\STATE $p_{stem} = \frac{1}{I_{stem}}$
% \STATE $p_{drop} = 1 - \frac{t}{T(I_{branch}+1)} if ** else $
\STATE $p_{branch} = \frac{1-p_{drop}}{I_{branch}} ~ if ~ o_i ~ \not= none ~ else ~ p_{drop}$
\STATE $E_{stem}^{'}$   = Sample($E_{stem}$, $p_{stem}$)
\STATE $E_{branch}^{'}$ = Sample($E_{branch}$, $p_{branch}$)
\STATE $m$              = $G(E_{stem}^{'} \cup E_{branch}^{'}, V)$
\STATE $D_{batch}$      = Sample($D_{train}$, $uniform$, $B$)
\STATE $\theta(m) = \theta(m) + \alpha \nabla L(\mathcal{F}(m, D_{batch})) + \sqrt{2\alpha }~\epsilon$
\IF{$t$ $\equiv$ $0$ $mod$ $T_{epoch}$}
\STATE $\mathcal{G}_{sample}$.Append($G(E,V)$)
\ENDIF
\ENDFOR
\STATE \textbf{Outputs: } $\mathcal{G}_{sample}$
\end{algorithmic}
\end{algorithm}

\subsubsection{Average Accuracy and Parameter}

Once we have $K$ samples of $\{\theta_T^{1},\theta_T^{2},...,\theta_T^{K}\}$ which approximates the parameters obtained by different run.
To eliminate the effect of random noise $\upsilon$ and stabilize the performance estimation, we propose two approaches: score expectation and parameter expectation.

Expectation over scores
approach is to define the score $S(m)$ of each model $m$ as the expectation of validation accuracy over $K$ sampled shared weights.
\begin{equation}
S_s(m) = \frac{1}{K}\sum_{i}^{K}Acc(m, \theta_i).
\end{equation}

Expectation over parameters approach is to take the average of sampled shared parameters and use average parameters to evaluate the performance of each model.
\begin{equation}
S_p(m) = Acc(m, \frac{1}{K}\sum_{i}^{K}\theta_i).
\end{equation}

\subsubsection{Independent fine-tuning}
When evaluating the single architecture performance, loads the weights from the hyper-network and resuming training the architecture independently should be able to get more architecture-relevant weights. Thus we test this approach in our experiment.

\subsection{Evolution Algorithm}
Inspired by recent work \cite{lu2018nsga,tan2019mnasnet}, we apply evolution algorithm NSGA-\uppercase\expandafter{\romannumeral2} as the search agent. In this section, we first introduce some basic concept of NSGA-\uppercase\expandafter{\romannumeral2}. Next we discuss how we apply NSGA-\uppercase\expandafter{\romannumeral2} to our search space.
\subsubsection{NSGA-\uppercase\expandafter{\romannumeral2}}
We seek to obtain the model architecture with excellent performance under the constraint of computational expense. NSGA-\uppercase\expandafter{\romannumeral2} is the most popular choice among multi-objective evolutionary method.
The core component of NSGA-\uppercase\expandafter{\romannumeral2},  is the Non Dominated Sorting which benefits the trade off between conflicting objectives. 
Since our optimization target is to minimize MAdds and maximize performance of architecture under different computational constraints.

% An architecture is encoded by number sequence where each item in the sequence represent a choice of certain block. We call the sequence architecture code. 
% To perform well on NSGA-\uppercase\expandafter{\romannumeral2} algorithm, the value of similar operation blocks should be closer to help building the trade-off frontier. 
% Totally, there are $43$ numbers in architecture code, where the first $19$ numbers includes the operation block of the sequential stem and the rest $24$ numbers stand for skip layers choice.

\subsubsection{Initialization}
To reduce manual bias and explore the search space better, we use random initialization for all individuals of the first generation. More specifically, each architecture randomly select basic operators for each block in the search space.
% 问题1：sampler 算novelty的话，写EA会很怪异，因为choice里没有None而只是超网训练采样时选了None

\subsubsection{Crossover and Mutation}
Single-point crossover on random position is adopted in our evolution algorithm. For two certain individuals $m_1 = (x_1, x_2, ..., x_n)$ and $m_2 = (y_1, y_2, ..., y_n)$, a single-point crossover strategy on position $p$ will result in a new individual $m_3=(x_1, x_2, ..., x_p, y_{p+1}, y_{p+2}, ..., y_n)$.

We use random choice mutation to enhance generation diversity. When a mutation happens to an individual, a selected operation block in it is changed to another available choice randomly.

% \begin{itemize}
%     \item $$ 1 - Accuracy $$
%     \item $$ MAdds $$
% \end{itemize}

\section{Experiments and Results}
We verify the effectiveness of our method on a large classification benchmark, ImageNet \cite{russakovsky2015imagenet}. In this section, we firstly describe our implementation details. Secondly, we present the performance of searching results on ImageNet as well as comparison with state of the art methods., Finally, we demonstrate the advantage of our designs via ablation study.

\subsection{Experiments Settings} \label{sec:exp-setting}

\textbf{Datasets} We conduct experiments on the ImageNet, a standard benchmark for classification task. It has $1.28M$ training images and $50K$ validation images. 
%We set the input resolution of the network to 224x224 and compare accuracy versus measures of multiply adds(MAdds).
%

\textbf{Train Details of Hyper-network} For the training of hyper-network, we adopt cosine learning rate scheduler with learning rate initialized as 0.1 and decaying to 2.5e-4 during $600$ epochs. A L2 regularization is used and its weight is set to 1e-4. The optimizer is mini-batch stochastic gradient decent (SGD) with batch size 512 and we set momentum as 0 to decouple the gradients of architectures sampled in different batches. Hyper-parameter $p_{drop}$ is set to $0.6$. The hyper-network is trained on 32 GTX-1080Ti GPUs. We implement the stabilized evaluation of our method by saving $20$ checkpoints at $600, 601, ..., 619$ epochs described in Sec. \ref{sec:mcmc}. The fine-tune strategy mentioned above is conducted with learning rate 2.5e-4.

\textbf{Search Details} The evolution agent randomly generates $45$ individuals for initialization. Then it repeats the exploitation and exploration loop where it generates $45$ individuals via single-point crossover and random mutation. It conducts $22$ loops and evaluates $990$ models. At last, we choose the top ranked 2 models under different MAdds constraints.

\begin{table*}[b]
\begin{center}
\begin{tabular}{l c c c c c}
\toprule[1.5pt]

Model & Search space & Params & MAdds & Top-1 acc & Top-5 acc\\
& &  (M) & (M) & (\%) & (\%) \\

\midrule[1pt]
DARTS \cite{liu2018darts} & Cell-based  & 4.7 & 574 & 73.3 & 91.3 \\
Proxyless-R \cite{cai2018proxylessnas} & Chain-structured  & - & 320 & 74.6 & 92.2 \\
Single-path NAS \cite{stamoulis2019single} & Chain-structured & 4.3 & 365 & 75.0 & 92.2 \\
FairNAS-A \cite{chu2019fairnas} & Chain-structured & 4.6 & 388 & 75.3 & 92.4 \\
FBNet-C \cite{wu2019fbnet} & Chain-structured & 5.5 & 375 & 74.9 & - \\

SPOS \cite{guo2019single}  & Chain-structured  & - & 328 & 74.7 & -\\
BetaNet-A \cite{fang2019betanas} & Chain-structured & 7.2 & 333 & 75.9 &  92.8 \\

\textbf{ST-NAS-A(ours)} & \textbf{Topology augmented} & \textbf{5.2} & \textbf{326} & \textbf{76.4} & \textbf{93.1} \\

% \midrule[1pt]

% \textbf{ST-NAS-B (ours)} & \textbf{Topology augmented} & \textbf{7.8} & \textbf{503} & \textbf{77.9}  & \textbf{93.8} \\

\bottomrule[1.5pt]
\end{tabular}
\end{center}
\caption{Performance comparison between ST-NAS and efficient NAS methods on ImageNet. Our model ST-NAS-A archieve best top 1 accuracy with the least MAdds.}
\label{tab:main-result}
\end{table*}

\textbf{Training Details of Resulted Architecture} For the independent training of resulted architectures, we use cosine learning rate scheduler with initial learning rate $0.8$. We train the model for 300 epochs with batch size 2048 and adopt SGD optimizer with nesterov and momentum value 0.9. To prevent overfitting, we use L2 regularization with weight 1e-4
and standard augmentations including random crop and colorjitter.

% After the evolution algorithm, we divide the results into several bins under the constraints. We choose the top-2 performance models of each bin and train them from scratch to get the results.
\begin{table*}[h]
\begin{center}
\begin{tabular}{l c c c c c}
\toprule[1.5pt]

Model & Search space & Params & MAdds & Top-1 acc & Top-5 acc\\
& &  (M) & (M) & (\%) & (\%) \\

\midrule[1pt]
% NSGA-NET(Lu \etal\cite{lu2018nsga}) &  &  &  &  &   \\ NO imagenet result
% PNAS-NET
% AmoebaNet(Regulized)

*MNASNet-A1 \cite{tan2019mnasnet} & Chain-structured  & 3.9 & 312 & 75.2 & 92.5 \\
*MNASNet-A2 \cite{tan2019mnasnet} &  Chain-structured & 4.8 & 340 & 75.6 & 92.7 \\
*RCNet-B \cite{xiong2019resource} & Chain-structured & 4.7 &  471 & 74.7 & 92.0 \\
*NASNet-B \cite{zoph2018learning}  & Cell-based & 5.3  & 488  & 72.8  & 91.3  \\
*EfficientNet-B0 \cite{tan2019efficientnet}) & Chain-structured  & 5.3 &  390 & 76.3 & 93.2 \\
\textbf{ST-NAS-A (ours)} & \textbf{Topology augmented} & \textbf{5.2} & \textbf{326} & \textbf{76.4} & \textbf{93.1} \\

\midrule[1pt]
% howard2019searching
1.4-MobileNetV2 \cite{sandler2018mobilenetv2} & Chain-structured & 6.9 & 585 & 74.7 & 92.5 \\
2.0-ShuffleNetV2 \cite{ma2018shufflenet} & Chain-structured & 7.4 & 591 & 74.9 & - \\
*NASNet-C \cite{zoph2018learning}  & Cell-based & 4.9  & 558  & 72.5  & 91.0  \\
*NASNet-A \cite{zoph2018learning}  & Cell-based & 5.3  & 564  & 74.0  & 91.6  \\
*1.4-MNASNet-A1 \cite{tan2019mnasnet} &  Chain-structured & - & 600 & 77.2 & 93.7 \\
*RENASNet \cite{chen2018reinforced}  & Cell-based & 5.4 & 580 & 75.7 & 92.6 \\
*PNASNet \cite{liu2018progressive} & Cell-based & 5.1 & 588 & 74.2 & 91.9 \\
\textbf{ST-NAS-B (ours)} & \textbf{Topology augmented} & \textbf{7.8} & \textbf{503} & \textbf{77.9}  & \textbf{93.8} \\

% MNASNet$\times 1.4+$SE & Chain-structured & & & \\

\bottomrule[1.5pt]
\end{tabular}
\end{center}
\caption{Performance comparison among ST-NAS, manual designed networks and sample based NAS methods on ImageNet. Notably sample based methods with mark $*$ takes much more computation resources. We show that the architecture discovered by ST-NAS perform better than both sample based NAS and manually designed architectures while maintaining least MAdds.}
\label{tab:sample-based}
\end{table*}

\subsection{Main Results} 
ST-NAS looks for models with objectives of low MAdds and high accuracy. We select two resulted models separately under small and large MAdds constraints, namely, ST-NAS-A and ST-NAS-B. Architectures and performance of them compared with state-of-the-art methods are discussed in this subsection.

\textbf{Performance on ImageNet.}
% Experiments results of our performance are shown in Table \ref{tab:main-result} and Table \ref{tab:sample-based}. The resulted architectures are shown in Fig. \ref{fig:result-net}.  Both ST-NAS-A and ST-NAS-B outperform state-of-the-art methods under comparable MAdds constraints. 
%\grh{ST-NAS-A performs better than MobileNetV2 with comparable MAdds by **.**\%. }
We compare ST-NAS method with efficient NAS methods, including DARTS, ProxyLessNAS and FBNet, in Table \ref{tab:main-result}. Our model ST-NAS-A outperforms all of them while with the least MAdds and comparable parameter number.

For architectures resulted from high cost, \ie, manually designed networks and networks obtained by sample-based methods, we compare ST-NAS with them in two groups divided by MAdds, as shown in Table \ref{tab:sample-based}. At a much less search cost, our ST-NAS outperform all the methods in both MAdds groups.

\begin{table}[h]
\begin{center}
\begin{tabular}{l | c c}
\toprule[1.5pt]

Model & MAdds (backbone) & mAP \\
 & (G) &  \\

\midrule[1pt]
MobileNetV2 & 0.33 &  31.7 \\
ST-NAS-A & 0.33 & 33.2 \\

\midrule[0.5pt]

ResNet18 & 1.81 & 32.2 \\
ST-NAS-B & 0.50 & 35.3 \\
ResNet50 & 4.09 & 36.9 \\
ST-NAS-B$^*$ & 1.03 & 37.7 \\
% ResNet101 & 7.9 & 39.4 \\

\bottomrule[1.5pt]
\end{tabular}
\end{center}
\caption{Performance on COCO dataset. The channel number of ST-NAS-B is scaled to get ST-NAS-B$^{*}$. ST-NAS-A outperforms MobileNetV2 by 1.5\% COCO AP while maintaining same MAdds. ST-NAS-B$^{*}$ achieves 0.5\% higher than ResNet50 but needs only a quarter MAdds.}
\label{tab:COCO-res}
\end{table}

\begin{figure*}[h]
\begin{center}
   \includegraphics[width=\linewidth]{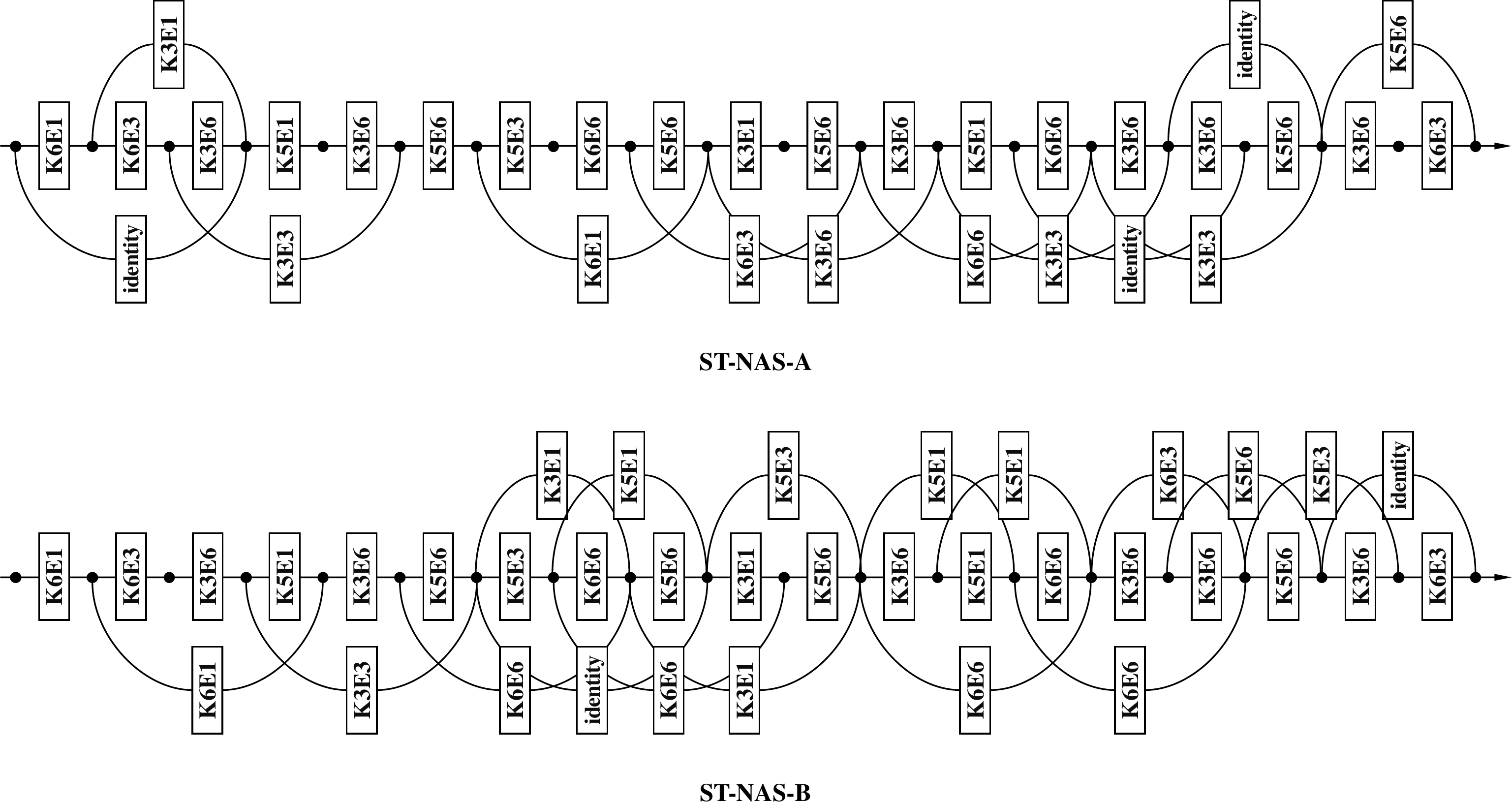}
\end{center}
\caption{Resulted architectures. Linear bottle-neck contains a $K\times K$ group-wise convolution layer between two $1\times 1$ point-wise convolution layers. Expand ratio is defined as the ratio between group-wise convolution channels and point-wise convolution channels. We describe a linear bottle-neck with its expand ratio, \ie, the number after ``E'', and its group-wise convolution kernel size, \ie, the number after ``K''. }% A linear bottle-neck contains a $K\times K$ group-wise convolution layer between two $1\times 1$ point-wise convolution layers.  The expand ratio is the ratio between group-wise convolution channels and point-wise convolution channels.}
%
% The most widely used blocks in our ST-NAS are \textbf{}linear bottle-neck The number after 'E' indicates the channel expand ratio of linear bottle-neck}
\label{fig:result-net}
\end{figure*}

\textbf{Performance on COCO.} Our implementation is based on feature pyramid network (FPN)\cite{lin2017feature}. Different models pretrained on ImageNet is utilized as feature extractor. All the models are trained for 13 epochs, known as $1\times$ schedule. %\cite{girshick2018detectron}.
The results are shown in Table \ref{tab:COCO-res}.
Our ST-NAS-A backbone outperforms MobileNetV2. The ST-NAS-B performs comparably with ResNet50 with much less MAdds.

\subsection{Ablation Studies}

\subsubsection{Rank Fluctuation}

To explain the importance of our stabilization mechanism, we randomly sample a set of architecture and rank their performance under shared parameters on validation set at the last 10 epochs training of hyper-network. As shown in Fig. \ref{fig:training_rank}, the rank of a single checkpoint fluctuates a lot during hyper-network training process and hardly distinguishes the performance of architectures, implying the necessity of a stabilized evaluation strategy.

% \begin{table}[]
% \begin{center}
% \begin{tabular}{l|cccc}
% \toprule[1.5pt]

%         & K=40 & K=20 & K=10 & K=5 \\
% \midrule[1pt]
% SGLD-acc $\tau$   & 0.817  & 0.81 &  0.79  & 0.68 \\
% SGLD-param $\tau$ & 0.83  & 0.82 &  0.77  & 0.66 \\

% \bottomrule[1.5pt]
% \end{tabular}
% \end{center}
% \caption{Ablation of checkpoint number K.}
% \label{tab:choosen-of-K}
% \end{table}

% \begin{table}[h]
% \begin{center}
% \begin{tabular}{ l c c}
% \hline 
% Evaluation & MAdds & Top-1 acc  \\
% \hline
% Baseline  &  &75.57  \\

% Fine-tune  &  &75.33 \\

% SGLD-param(Constant)  & & 75.44 \\

% SGLD-acc(Constant) & & 75.47 \\

% SGLD-param(Vary) & & xx.xx \\

% SGLD-acc(Vary) & & xx.xx \\

% \hline
% \end{tabular}
% \end{center}
% \caption{Results of different evaluation strategy. "Baseline" means direct use single checkpoint. 'Vary' means a varied step length is used in SGLD.}
% \label{evaluation_ab}
% \end{table}
\begin{table}[b]
\begin{center}
\begin{tabular}{l c}
\toprule[1.5pt]
Estimation approach  & $ \tau $\\

\midrule[1pt]

Single checkpoint   &  0.71  \\

Fine-tune           &  0.64  \\

SGLD-param          &  0.84  \\

SGLD-acc            &  0.81  \\
\bottomrule[1.5pt]
\end{tabular}
\end{center}
\caption{$\tau$ of different rank stabilization approach we proposed. The original baseline is single checkpoint achieves 0.71 of Kendall $\tau$ which is 0.07 higher than the fine-tune approach. The parameter expectation and accuracy expectation method is tied and outperform the baseline with a margin.}
\label{tab:tau-diff-method}
\end{table}

% \subsubsection{Evaluation Strategy}

% As mentioned in section \ref{approach}, we propose fine-tune, SGLD with average accuracy and SGLD with average parameter to evaluate the performance of architecture. To compare the accuracy of model ranking of these strategies, we show the performance of resulted architecture of separately using one of the three strategies in our evaluation stage. As shown in Table \ref{tab:tau-diff-method}, SGLD outperforms fine-tune due to its reduction of the variance of parameter distribution. 

% \begin{table}[h]
% \begin{center}
% \begin{tabular}{l c}
% \toprule[1.5pt]
% Estimation approach  & Top-1 acc(\%) \\

% \midrule[1pt]
% % 数据来源于600epoch + cosine fine-tune

% Single checkpoint   &  77.6  \\

% Fine-tune   &   75.8   \\

% SGLD-param    & 77.9   \\

% SGLD-acc  & 77.6   \\
% \bottomrule[1.5pt]
% \end{tabular}
% \end{center}
% \caption{Model accuracy comparison of different evaluation strategy.}
% \label{tab:acc-diff-method}
% \end{table}

\subsubsection{Ranking Verification}

% \begin{figure}[h]
% \begin{center}
%   \includegraphics[width=0.95\linewidth]{pic/sensibility-K.pdf}
% \end{center}
% \caption{Sensibility of SGLD w.r.t. the number of samples $K$. The correlation with ground truth of both the two SGLD variants increases with K, as more accurate expectation is achieved. SGLD-Param outperforms SGLD-Acc at large $K$s, because SGLD-Acc uses multiple parameters, all of which are noisy, while SGLD-Param uses a less noisy parameter by estimation of expectation.}
% \label{fig:choosen-of-K}
% \end{figure}

% \begin{table}[h]
% \label{tab:unknown}
% \begin{center}
% \caption{Ranking ability of evaluation strategy. "Baseline" means direct use single checkpoint. 'Vary' means a varied step length is used in SGLD.}

% \begin{tabular}{ l c}
% \hline 
% Evaluation  &Tau \\
% \hline
% Baseline    &0.55  \\
% %& & & \\
% \hline
% Fine-tune   &0.60(early)   \\
% %& & & \\
% \hline
% SGLD-param(Constant)    &0.73   \\
% %& & &\\
% \hline
% SGLD-acc(Constant)  &0.73   \\
% \hline
% SGLD-param(Vary)    &0.73   \\
% %& & &\\
% \hline
% SGLD-acc(Vary)  &0.73   \\
% %& & &\\
% \hline
% \end{tabular}
% \end{center}
% \end{table}

We further verify this reduction by quantifying the ranking ability of different evaluation strategies by correlation coefficient between ranks in hyper-network and the ground truth ranks. Kendall's tau coefficient is adopted as the metric in our verification. We randomly sample 12 networks and train them form scratch to obtain the ground truth rank. To compare with single checkpoint, we make use of  checkpoints of 10 epochs at the 591-th, 592-th, ..., 600-th epoch to generate 10 ranks and get 10 correlation coefficients with the ground truth rank. The median of the 10 correlation coefficients is adopted to compare with other strategies. It is observed in Table \ref{tab:tau-diff-method} that SGLD consistently achieves higher correlation coefficients than fine-tune and single checkpoint, which verifies the effectiveness of SGLD in the reduction of parameter variance.

\section{Conclusion}
We proposed a topology-diverse search space and a novel search method, ST-NAS. In ST-NAS, we improve both the sampling strategy during hyper-network training and the architecture evaluation approach by rigorous theoretical analysis. Sound experiments demonstrate the effectiveness of our designs and achieve consistent improvements under different computation cost constraints.

% \clearpage
\bibliographystyle{splncs04}
\bibliography{eccv20_conference}
\end{document}